\documentclass[10pt, conference]{IEEEtran}
\usepackage{graphicx}
\usepackage[backend = bibtex,
			style = numeric,
			sorting = none,
			]{biblatex}
			
{\footnotesize
\bibliography{references.bib}}
\usepackage{comment}
\usepackage{fixltx2e}
\usepackage{amsmath}
\usepackage{varwidth}
\usepackage[dvipsnames]{xcolor}
\usepackage{subfigure}
\usepackage{enumitem}
\usepackage{cleveref}
\usepackage{caption}
\usepackage{multirow}

\newcommand{\eg}[1]{}
\renewcommand{\eg}[1]{e.g. {#1}}

\newcommand{\ie}[1]{}
\renewcommand{\ie}[1]{i.e. {#1}}

\newcommand*{\NODEBUG}{}

\ifdefined\DEBUG
\usepackage{todonotes}
\newcommand{\anote}[1]{\textcolor{purple}{[AG:#1]}}
\newcommand{\anote}[1]{}
\newcommand{\gnote}[1]{\textcolor{magenta}{[GKT:#1]}}
\newcommand{\jnote}[1]{\textcolor{olive}{[JCD:#1]}}
\newcommand{\ynote}[1]{\textcolor{red}{[YHL:#1]}}
\newcommand{\cnote}[1]{\textcolor{blue}{[CT:#1]}}
\newcommand{\xnote}[1]{\textcolor{royal}{[XH:#1]}}
\newcommand{\hnote}[1]{\textcolor{green}{[HW:#1]}}
\fi

\ifdefined\NODEBUG
\usepackage[disable]{todonotes}
\newcommand{\anote}[1]{}
\newcommand{\gnote}[1]{}
\newcommand{\jnote}[1]{}
\newcommand{\ynote}[1]{}
\newcommand{\cnote}[1]{}
\newcommand{\xnote}[1]{}
\newcommand{\hnote}[1]{}
\newcommand{\nnote}[1]{}
\newcommand{\delete}[1]{}
\fi

\IEEEoverridecommandlockouts
\IEEEpubid{\makebox[\columnwidth]{978-1-6654-3922-0/21/\$31.00 $\copyright$2021 IEEE \hfill} \hspace{\columnsep}\makebox[\columnwidth]{ }}

\begin{document}

\title{Low-Power Multi-Camera Object Re-Identification using Hierarchical Neural Networks}

\author{\IEEEauthorblockN{Abhinav Goel,
Caleb Tung, Xiao Hu, Haobo Wang,
James C. Davis, George K. Thiruvathukal\IEEEauthorrefmark{1}, Yung-Hsiang Lu}
\IEEEauthorblockA{Purdue University, School of Electrical and Computer Engineering, West Lafayette, IN, USA \\
\IEEEauthorrefmark{1}Loyola University Chicago, Department of Computer Science, Chicago, IL, USA}}

\maketitle

\begin{abstract}
\jnote{Do we need this first sentence?}
Low-power computer vision on embedded devices has many applications. 
This paper describes a low-power technique for the object re-identification (reID) problem: matching a query image against a gallery of previously-seen images.
State-of-the-art techniques 
rely on large, computationally-intensive Deep Neural Networks (DNNs).
We propose a novel hierarchical DNN architecture that uses attribute labels in the training dataset to perform efficient object reID.
At each node in the hierarchy, a small DNN identifies a different attribute of the query image.
The small DNN at each leaf node is specialized to re-identify a subset of the gallery---only the images with the attributes identified along the path from the root to a leaf. 
Thus, a query image is re-identified accurately after processing with a few small DNNs. 
We compare our method with state-of-the-art object reID techniques.
With a $\sim$4\% loss in accuracy, our approach realizes significant resource savings: 74\% less memory, 72\% fewer operations, and 67\% lower query latency, yielding 65\% less energy consumption. 

\end{abstract}

\maketitle

\section{Introduction}
\vspace{-0.05in}
\label{sec:intro}

Object re-identification (reID) is an important computer vision task. Given an image of an object, the goal is to return a group of images containing the queried object~\cite{pyramid}. Systems employing object reID can be used to enhance public safety~\cite{covid}, manage crowds~\cite{sara, covid-1}, and detect events across multiple cameras~\cite{sem1}. In these applications, object reID systems may be deployed on embedded devices like traffic cameras and drones~\cite{autoreid}. 
On such devices, computing resources are scarce and energy efficiency is critical. 
Existing object reID systems are not suitable for embedded devices because they require large Deep Neural Networks (DNNs) that are compute and memory intensive~\cite{goel2020survey}.
Large DNNs are used for object reID because images are captured from multiple non-overlapping cameras with varying angles, backgrounds, and~occlusions~\cite{pyramid}.



\begin{figure}
    \centering
    \includegraphics[width=0.95\linewidth]{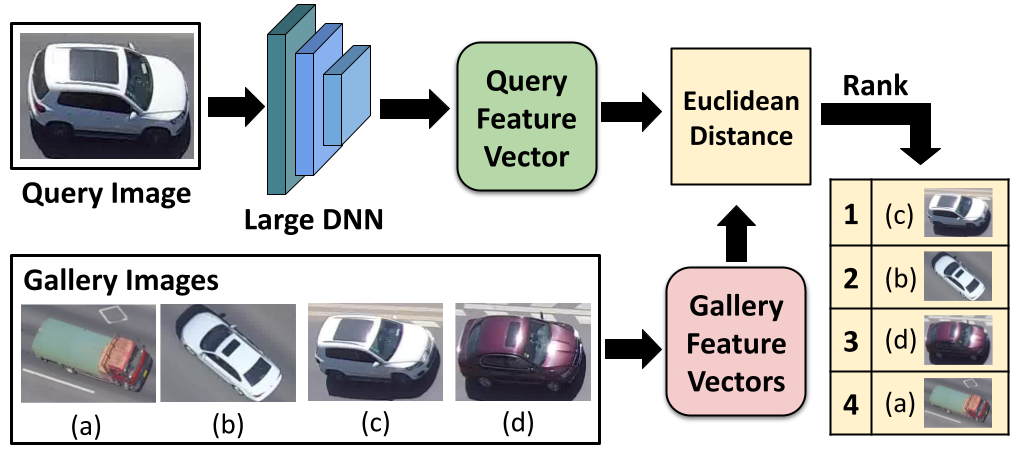}
    \caption{
    Existing object reID systems use large DNNs to extract a feature vector from a query image to check if the system has seen the query before in the gallery. The system compares the query feature vector with extracted feature vectors for \emph{every} gallery image, incurring large computation expense.}

    
    \label{fig:reid}
    \vspace{-0.25in}
\end{figure}

The typical working of existing object reID techniques~\cite{Densenet} is depicted in Fig.~\ref{fig:reid}. A large DNN extracts a feature vector from the query image. This feature vector is then compared with the feature vectors of every gallery image, \ie images (a) - (d), using a distance metric such as the Euclidean Distance. The gallery images are then ranked based on their distance from the query image. The existing techniques perform many redundant operations because query images need not be compared with every gallery image.
For example, the query image in Fig.~\ref{fig:reid} (a white car with a sunroof) could be compared only to other white cars with sunroofs --- gallery images (b) and (c). 

This paper proposes to use a hierarchical DNN to identify objects with the same attributes (\eg color, body style, sunroof, etc.) for high accuracy and fewer redundant operations. A hierarchical DNN is an architecture that uses several DNNs in the form of a hierarchy~\cite{treecnn}. Although hierarchical DNNs have been used to reduce redundant operations in computer vision~\cite{islped}, there are challenges associated with using them for efficient object reID (Section~\ref{sec:back}). This paper employs the hierarchical DNN architecture, with innovations in the similarity metrics and the method to define the hierarchy structure to adapt it to the object reID problem.





\begin{figure}[t!]
    \centering
    \includegraphics[width=\linewidth]{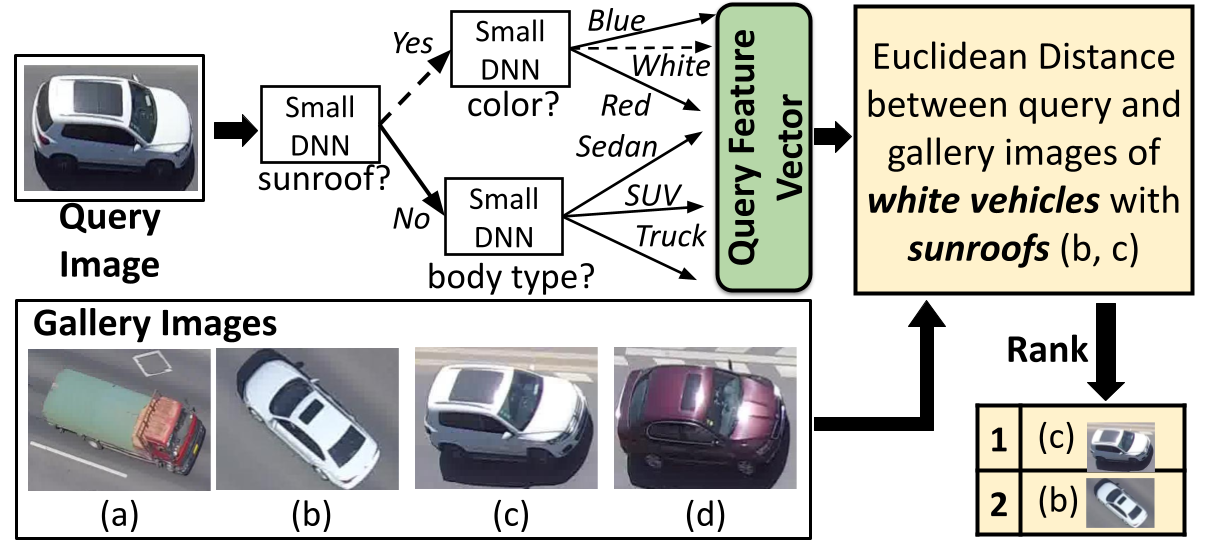}
    \caption{
    Illustration of the proposed method: A query image is processed by a hierarchical DNN.
    Each small DNN extracts a feature vector and identifies an attribute.
    Each query image follows one path from the root to a leaf (dashed line).
    This technique only computes distances between, and ranks, gallery images with the same attributes.
    }
    \label{fig:hier}
    \vspace{-0.2in}
\end{figure}

In our architecture, each node of the hierarchy contains a small DNN that extracts a feature vector from the query and routes the query among its subsequent branches.
Fig.~\ref{fig:hier} illustrates our approach. The query image is processed by the first small DNN to obtain a feature vector. This DNN determines if the vehicle has a \textit{sunroof}. After the query image is classified as a vehicle with a sunroof by the first DNN, the gallery reduces to images (b), (c), and (d). The next DNN continues to process the feature vector and identifies the vehicle's \textit{color}. This classification reduces the gallery to images (b) and (c). This process continues until a leaf node is reached. The feature vector from the leaf DNN is used to perform comparisons with the remaining gallery images to re-identify the object. The remaining gallery images contain objects with attributes identified in the query.
Because each small DNN specializes in processing only a subset of the objects (with specific attributes), they obtain high accuracy. 



This paper considers two object reID applications during experiments: vehicle reID~\cite{VRAI} and person reID~\cite{Market}.
With a $\sim$4\%, loss in rank-1 accuracy, we show that the proposed method substantially reduces resource requirements. We observe reductions in memory requirement by 74\%-97\%, query latency by 67\%-89\%, operations by 72\%-93\%, and energy by 65\%-88\%. The experiments are conducted on two embedded devices: Raspberry Pi 3 and NVIDIA Jetson Nano.




\section{Background and Related Work}
\vspace{-0.02in}

\subsection{Hierarchical Deep Neural Networks}
\label{sec:back}

Hierarchical DNNs use multiple DNNs in a tree structure~\cite{todaes, islped}.
An input is characterized by the path it follows from root to leaf (Fig.~\ref{fig:hier}). 
Recently these techniques have been used to perform classifications at every level of the tree to reduce the problem size and improve efficiency~\cite{islped}.



Hierarchical DNN architectures are constructed using either visual~\cite{treecnn, islped} or semantic~\cite{Yolo9000} similarities. These techniques, however, are not applicable to object reID.
Roy et al.~\cite{treecnn} show that hierarchies based on visual similarities  need to be re-trained whenever a previously unseen object is encountered.
Meanwhile, hierarchies based on semantic similarities have significant accuracy losses~\cite{islped} when used for computer vision.
We address these shortcomings with a novel combination of visual and semantic similarities, and show that this combination is advantageous for efficient object reID.



\vspace{-0.05in}
\subsection{Related Work}


\textbf{Object Re-Identification: }
In object reID, the task is to identify if an object exists in a gallery (database) of objects that have appeared before.
There are two popular methods to perform object reID: (1)~\textit{Global feature vectors}: use DNNs to obtain a large feature vector for each image~\cite{wang2018, DGNet}. (2)~\textit{Local feature vectors}: combine multiple smaller feature vectors (\eg different body/vehicle parts) to form a single feature vector for each image~\cite{pyramid, autoreid, PCB, RAM}. In addition to feature vectors, some techniques use \textit{auxiliary information} (e.g. attributes or synthetic images) to improve reID accuracy~\cite{sem1, VRAI}.
Techniques using auxiliary feature vectors often require datasets annotated with attribute labels~\cite{sem1, VRAI}.
Our method uses global features and auxiliary information to construct a hierarchy of small DNNs for low-power object reID on embedded devices.



\textbf{Low-Power Computer Vision:}
Goel et al.~\cite{goel2020survey} survey low-power DNNs and describe the benefits of reducing memory and operations for low-power applications.
DNN quantization reduces the memory requirement~\cite{Goel2018} and DNN pruning reduces the DNN operations~\cite{lpirc}. Although these techniques increase the efficiency of existing large DNNs, they generally lower the accuracy as well.
In contrast, our method uses hierarchical DNNs to reduce the power consumption of object reID. To the best of our knowledge, there are no existing techniques for low-power object reID on embedded devices.



\textbf{Contributions:}
(1)~This is the first method to use hierarchical DNNs to reduce the resource requirements of object reID for low-power embedded devices.
(2)~The proposed method presents a novel technique that uses both semantic and visual similarities to construct hierarchies for object reID.  
(3)~Our hierarchical approach achieves high accuracy with low-resource requirements by using small DNNs that are specialized in processing a small subset of the gallery.
\jnote{Consider cutting the next sentence, it's not really a contribution per se. You already said this stuff at the end of the introduction (just half a page ago!) in more detail.}
Experiments show that our method outperforms existing techniques in terms of memory, number of operations, and energy. 

\section{Hierarchical Object Re-identification}
\vspace{-0.05in}


This section describes the proposed hierarchical object reID technique. Section~\ref{sub:order} describes the method to define the hierarchy structure. Section~\ref{sub:cons} describes how to construct the DNNs for each node of the tree. Section~\ref{sec:train} discusses the training process. Section~\ref{sec:reID} explains how object reID is performed with the proposed method. We use examples from Market-1501~\cite{Market} to explain the proposed techniques.

\begin{figure}[b!]
    \vspace{-0.15in}
    \centering
    \includegraphics[height = 1.3in]{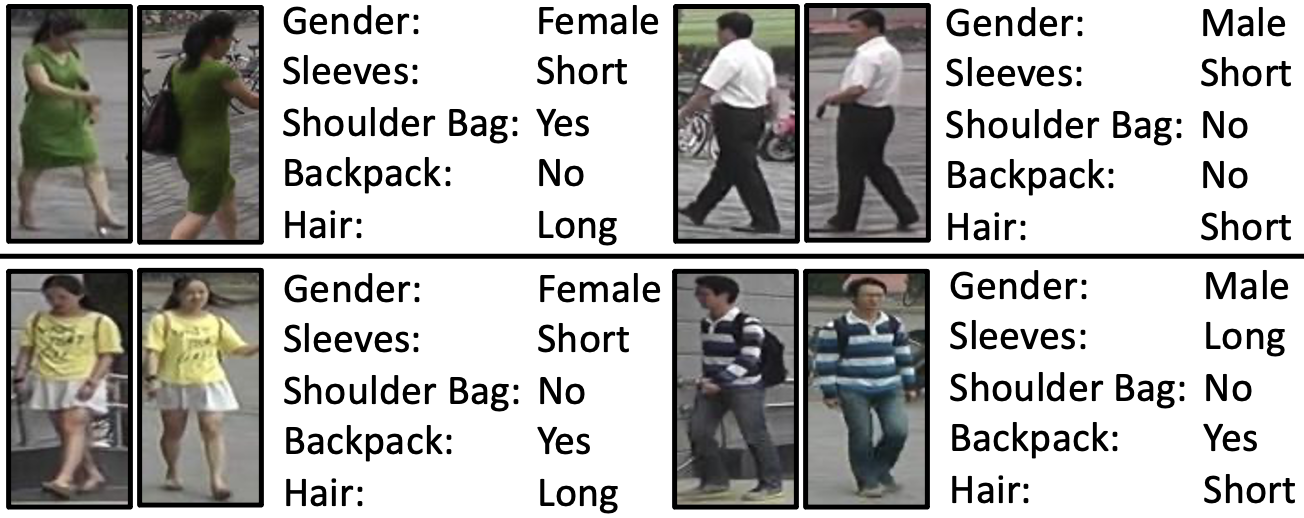}
    \caption{
    Examples of images in the Market-1501 dataset with some of their attributes. When defining the structure of the hierarchical DNNs, we must determine which attributes should be identified and the order in which they are identified.
    }
    \label{fig:atts}
    \vspace{-0.05in}
\end{figure}

\begin{figure*}[t!]
        \centering
        \subfigure[]{\includegraphics[height = 1.26in, width = 1.34in]{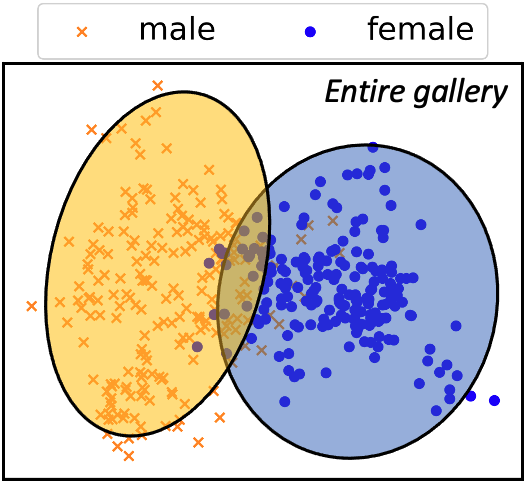}}
        \subfigure[]{\includegraphics[height = 1.26in, width = 1.34in]{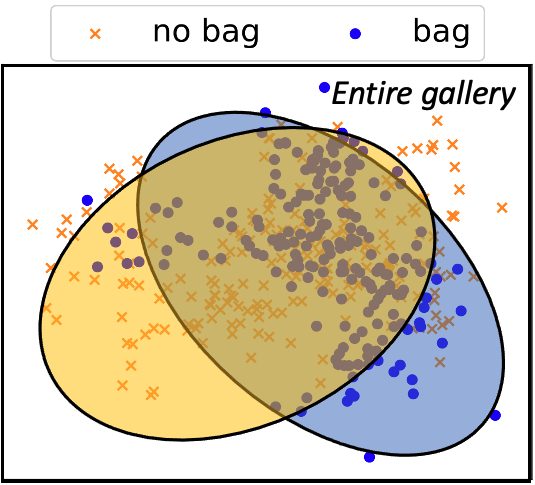}}
        \subfigure[]{\includegraphics[height = 1.26in, width = 1.34in]{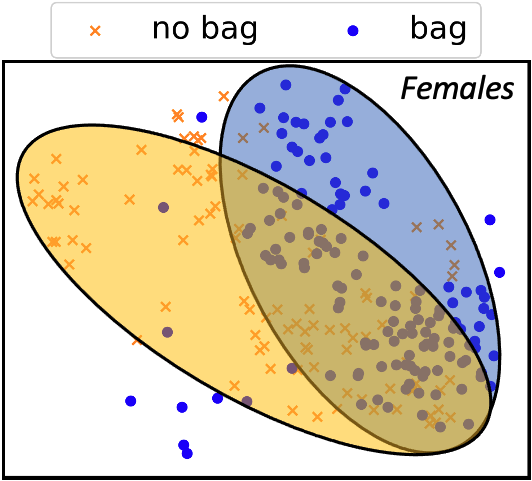}}
        \subfigure[]{\includegraphics[height = 1.26in, width = 1.34in]{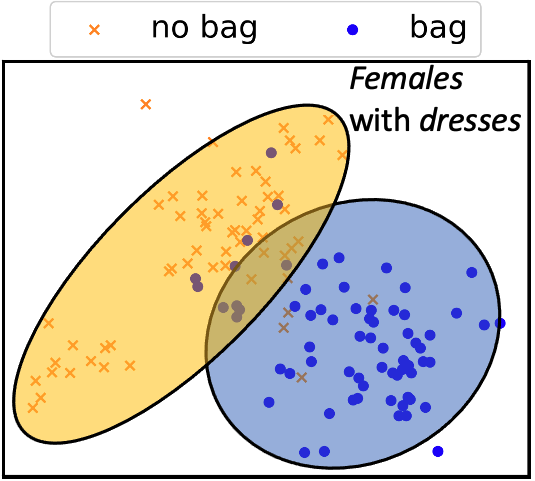}}
        \subfigure[]{\includegraphics[height = 1.26in]{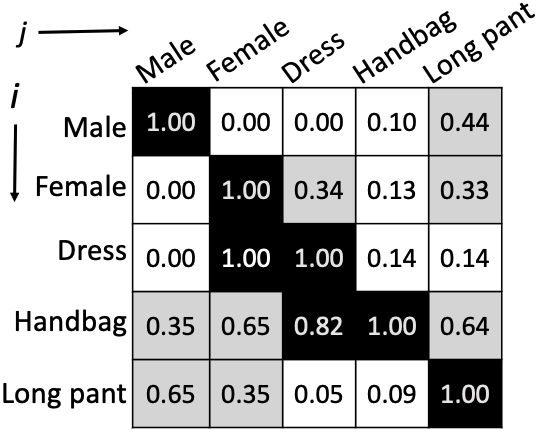}}
         \vspace{-0.1in}
        \caption{(a), (b) Feature vectors (after principal component analysis) of images from the entire Market-1501 training gallery~\cite{Market} and labeled by (a) \textit{gender}, (b) \textit{bag-carrying}. \textit{Gender} is easier to identify than \textit{bag} because it has more discernible clusters. (c),~(d)~Feature vectors of subsets of the Market-1501 training gallery, (c)~only \textit{females}, (d)~only \textit{females} wearing \textit{dresses}. The difficult \textit{bag} attribute identification becomes progressively easier (clusters are more discernible) on smaller subsets of the gallery. (e)~Excerpt of the correlation matrix for Market-1501, showing the likelihood that a person has attribute $j$ if they have attribute $i$. Black boxes: high positive correlation. White boxes: high inverse correlation. Grey boxes: low correlation. Attributes with low correlation with the identified attributes are identified deeper in the hierarchy.}
        \label{fig:clusters}
        \vspace{-0.2in}
\end{figure*}

\vspace{-0.05in}
\subsection{Defining the Hierarchy Structure}
\label{sub:order}
\jnote{Given the taxonomy you present in the Background, I think we should be more explicit about the auxiliary information we depend on. It looks to me like we need (a) labels, and (b) an existing pre-trained DNN to measure attribute difficulty. (Later, it would be nice if we can squeeze in thoughts about `...and this is what limitations that might introduce').}

The proposed technique uses multiple small DNNs that extract feature vectors from the query image and identify attributes to reduce the size of the gallery. To construct the hierarchy, our technique determines (1)~\emph{which} attributes are identified and (2)~the \textit{order} in which they are identified. We provide the intuition of our approach using Fig.~\ref{fig:atts} and~\ref{fig:clusters}.


Fig.~\ref{fig:atts} shows examples of images from the Market-1501 dataset with some of their attributes. Identifying all the attributes in images is unnecessary. For example, if most people in the database who are carrying \textit{backpacks} are not also carrying \textit{shoulder bags}, then identifying if a person is carrying a \textit{shoulder bag} is likely to be redundant if a \textit{backpack} is identified. 
Our technique uses attribute correlations to determine which attributes should be identified. Correlations are the likelihood of finding two attributes on the same object; highly correlated attributes need not be identified in the same path because the expected gallery size reduction is small. 

Fig.~\ref{fig:clusters}~(a, b) show feature vectors obtained using a large pre-trained DNN~\cite{Densenet} for the images in the Market-1501 dataset. Contours are added to the clusters to highlight the varying difficulty of attribute identifications. Attributes with clearer clusters (\eg{\textit{gender}}) are easier to identify than others (\eg{\textit{bag-carrying}}). 
In the Market-1501 dataset, males and females are visually dissimilar, but people with and without bags are difficult to distinguish.
In consequence, \textit{gender} classification should be performed close to the root of the hierarchy. This is because a small DNN can identify visually dissimilar attributes accurately.
Furthermore, difficult classifications become easier closer to the leaves of the hierarchy, because the classifications are performed on subsets of the gallery.
Fig.~\ref{fig:clusters} depicts the difficulty of identifying \textit{bags} when the gallery contains (b)~all images, (c)~only~\textit{females}, and (d)~\textit{females} wearing \textit{dresses}, respectively. The difficult \textit{bag} classification becomes progressively easier when performed on smaller subsets of the gallery. 
To ensure small DNNs can perform object reID accurately, we need to determine the \emph{order} in which attributes are identified.

In this paper, we use the difficulty of classifications (visual similarity)
and find attribute correlations (semantic similarity) to determine which attributes are identified and in what order.


\subsubsection{Quantifying the Difficulty of Attribute Identification} 
\label{subsub:difficulty}
To determine which attribute is identified at the root, and the order of the subsequent attribute classifiers in the hierarchy, we quantify the difficulty of attribute identification.
We perform easier attribute identifications close to the root of the hierarchy.
This ordering is advantageous because hierarchical DNNs propagate the output feature vectors from parent to child.
Each branch of a hierarchical DNN represents a deeper DNN (\ie more layers). 
The small DNNs near the leaves have greater distinguishing capabilities than the DNNs close to the root. 

To measure the difficulty of the attribute classifications, we use the Linear Evaluation Protocol~\cite{kent}. A linear classifier is trained on the feature vectors of a pre-trained DNN, and the validation error is obtained to measure the difficulty of the classification.  
Attributes with larger validation errors are more difficult to classify.
Sorting attributes by their validation error allows us to create a rank list of attributes based on their difficulty. 
For the Market-1501 dataset, \textit{gender} is the top-ranked attribute with linear classification error~$= 0.08$, and \textit{bag} is the bottom-ranked attribute with linear classification error~$= 0.23$. The top-ranked attribute is identified at the root of the tree.  The subsequent attribute identifications for each branch of the tree are determined by using a combination of the difficulty of identification and the attribute correlations. 

\subsubsection{Quantifying the Attribute Correlations}
\label{subsub:corr}
The first attribute classification (\ie the top-ranked attribute) of the hierarchy is determined after quantifying the difficulty of the attribute identifications. 
Then, for each branch of the tree, we need to determine which attributes to classify. This is done by recursively obtaining the correlations between different attributes; highly (positively or inversely) correlated attributes are not identified in the same branch. 
A correlation matrix $C_{i,j}$, in Fig.~\ref{fig:clusters}(e), shows the correlation between pairs of attributes: the likelihood of a person having attribute $j$, if the same person also has attribute $i$: $\text{Pr}(j \vert i)$. 
Each entry of $C_{i,j}$ is obtained by dividing the number of training dataset images containing the attributes $i$ and $j$ by the number of training dataset images containing attribute $i$.
A large  $C_{i,j}$ indicates a high positive correlation between attributes $i$ and $j$ (\eg{$C$\textsubscript{dress, female}$= 1.00$}).
A small $C_{i,j}$ indicates $i$ and $j$ have high inverse correlation (\eg{$C$\textsubscript{dress, male} $= 0.00$}). 
The correlation matrix is computed at each node of the tree, for every attribute $k$ with respect to the set of attributes $i$ to $j$, that have been already been identified along the path from the root to the node: $C_{i...j,k} = \text{Pr}(k | i...j)$.

The value of the correlation matrix can be understood with an example. Following the \textit{gender} classification, determining if the person is wearing a \textit{dress} is only useful if the person is female ($C$\textsubscript{female, dress} $= 0.34$). Identifying a \textit{dress} is not useful for males ($C$\textsubscript{male, dress} $= 0.00$). The \textit{dress} classification output can be inferred from the high inverse correlation between males and dresses. 
Performing the \textit{dress} classification for males does not help reduce the gallery size, and thus is a redundant operation.

Identifying attributes that have a low correlation to the previously identified attributes reduces redundant operations. 
If there are multiple unidentified attributes with low correlation, the attribute with the highest rank (easiest to identify), as determined in Section~\ref{subsub:difficulty}, is selected as the next attribute classification.
If there are no unidentified attributes with low correlation, no more attribute classifications are performed (leaf node).
To operationalize this, we consider attributes with $C_{i,j} \in [0.3, 0.7]$ as weakly correlated.


The proposed method increases efficiency for the common case by only identifying attributes that are weakly correlated to the previously identified attributes. Consider an example: in Market-1501, most males do not carry \textit{handbags}  ($C$\textsubscript{male, handbag} $= 0.10$). 
For re-identifying all males (with or without \textit{handbags}), other attributes (\eg{ \textit{age}, \textit{bag}, etc.}) are identified to reduce the gallery size. 
For the common case, \ie males without \textit{handbags}, fewer redundant operations are performed. 

\vspace{-0.05in}
\subsection{Constructing the Hierarchy}
\label{sub:cons}


Once the structure of the hierarchy is defined,
DNNs are constructed for each node.
Each DNN performs two tasks: (1)~extract feature vectors, and (2)~identify attributes.

Each node of the hierarchy contains a small DNN, that specializes in processing and re-identifying a subset of the gallery.
The DNN architectures need to be selected such that each DNN can perform the tasks accurately and efficiently. Larger DNNs are more accurate but use more resources. To obtain an acceptable tradeoff between accuracy and efficiency, we apply an architecture search technique~\cite{todaes}. It uses the change in accuracy density to evaluate whether a DNN ($D_{i+1}$) with $i+1$ layers 
should be selected over a DNN ($D_i$) with $i$ layers. The change is represented as $\Delta AD(D_i, D_{i+1}) = \frac{a_{i+1} - a_i}{m_{i+1}- m_i}$, where $a_i$ and $m_i$ are the accuracy and the memory requirement of $D_i$, respectively. 
This technique increases the DNN size one layer at a time and computes $\Delta AD(D_i, D_{i+1})$ until there are diminishing returns in accuracy with larger models.
In our case, we select $a_i$ as the average of the reID mean average precision and the attribute classification accuracy. 

Each small DNN follows the structure of the dense block from DenseNet~\cite{Densenet}, because of its ability to extract informative feature vectors.
The root of the hierarchy uses a $7 \times 7 \times 64$ convolution kernel with stride 2 and a maxpool $2 \times 2$ layer to downsample the image. Each subsequent layer is a ``dense layer'', \ie a layer containing a sequence of $1 \times 1 \times 128$ and $3 \times 3 \times 32$ operations. The number of subsequent dense layers is determined by the architecture search method. Average-pooling is used to resize the tensors before extracting a $128 \times 1$ feature vector and the classification output. The activation map before the average-pool layer is used as input to the chosen child DNN. Each child DNN follows the same structure of dense layers. TABLE~\ref{tab:arch} shows some of the DNNs obtained with this method.



\vspace{-0.05in}
\subsection{Training Method}
\label{sec:train}

The DNNs of the hierarchy are trained using backpropagation. We use two loss functions: (1)~triplet loss with batch hard mining for training DNNs to extract a feature vector such that the feature vectors of the same object are similar~\cite{wang2018}. (2)~cross-entropy loss to perform attribute classifications~\cite{resnet}.

The two losses are used separately because they require different training batch configurations.
First, the DNN is trained with triplet loss to extract a feature vector.
Then, the feature vector extraction parameters are frozen (not updated via backpropagation) and the subsequent classification layers are trained with the cross-entropy loss function. 
The DNNs of the hierarchy are trained in a root-down fashion: the root DNN is trained first, then its children, and so on. When training a DNN, the parameters of the ancestor DNNs are frozen. 




\begin{table}
    \vspace{0.1in}
    \centering
    \begin{tabular}{lllr|lr|}
    \hline
    \multicolumn{2}{|c|}{Root (gender)} &
      \multicolumn{2}{c|}{Child 1 (length of pant)} &
      \multicolumn{2}{c|}{Child 2 (Age)} \\ \hline
    \multicolumn{2}{|l|}{\begin{tabular}[c]{@{}l@{}}conv7$\times$7$\times$64\\ maxpool2$\times$2\end{tabular}} &
      \multicolumn{2}{l|}{maxpool2$\times$2} &
      \multicolumn{2}{l|}{maxpool2$\times$2} \\ \hline
    \multicolumn{2}{|l|}{\begin{tabular}[c]{@{}l@{}}conv1$\times$1$\times$128\\ conv3$\times$3$\times$32\end{tabular}} &
      \multicolumn{2}{l|}{\begin{tabular}[c]{@{}l@{}}conv1$\times$1$\times$128\\ conv3$\times$3$\times$32\end{tabular}} &
      \multicolumn{2}{l|}{\begin{tabular}[c]{@{}l@{}}conv1$\times$1$\times$128\\ conv3$\times$3$\times$32\end{tabular}} \\ \hline
    \multicolumn{2}{|l|}{\begin{tabular}[c]{@{}l@{}}conv1$\times$1$\times$128\\ conv3$\times$3$\times$32\end{tabular}} &
      \multicolumn{2}{l|}{\begin{tabular}[c]{@{}l@{}}conv1$\times$1$\times$128\\ conv3$\times$3$\times$32\end{tabular}} &
      \multicolumn{2}{l|}{\begin{tabular}[c]{@{}l@{}}conv1$\times$1$\times$128\\ conv3$\times$3$\times$32\end{tabular}} \\ \hline
    \multicolumn{2}{|l|}{\begin{tabular}[c]{@{}l@{}}conv1$\times$1$\times$128\\ conv3$\times$3$\times$32\end{tabular}} &
      \multicolumn{2}{l|}{\begin{tabular}[c]{@{}l@{}}conv1$\times$1$\times$128\\ conv3$\times$3$\times$32\end{tabular}} &
      \multicolumn{2}{l|}{\begin{tabular}[c]{@{}l@{}}conv1$\times$1$\times$128\\ conv3$\times$3$\times$32\end{tabular}} \\ \hline
    \multicolumn{2}{|l|}{avgpool} &
      \multicolumn{2}{l|}{avgpool} &
      \multicolumn{2}{l|}{\begin{tabular}[c]{@{}l@{}}conv1$\times$1$\times$128\\ conv3$\times$3$\times$32\end{tabular}} \\ \hline
    \multicolumn{1}{|c}{\begin{tabular}[c]{@{}c@{}}Output1: \\ Output2:\end{tabular}} &
      \multicolumn{1}{r|}{\begin{tabular}[c]{@{}r@{}}128$\times$1\\ 2$\times$1\end{tabular}} &
      \multicolumn{1}{c}{\begin{tabular}[c]{@{}c@{}}Output1: \\ Output2:\end{tabular}} &
      \begin{tabular}[c]{@{}r@{}}128$\times$1\\ 2$\times$1\end{tabular} &
      \multicolumn{2}{l|}{avgpool} \\ \hline
     &
       &
      \multicolumn{1}{c}{} &
       &
      \multicolumn{1}{c}{\begin{tabular}[c]{@{}c@{}}Output1: \\ Output2:\end{tabular}} &
      \begin{tabular}[c]{@{}r@{}}128$\times$1\\ 4$\times$1\end{tabular} \\ \cline{5-6} 
    \end{tabular}

    \caption{The DNNs constructed for Market-1501 dataset achieve an acceptable tradeoff between accuracy and efficiency. The root DNN contains 3 dense layers.}
    \vspace{-0.2in}
    \label{tab:arch}
\end{table}

\begin{figure}[b!]
    \vspace{-0.2in}
    \centering
    \includegraphics[width=0.96\linewidth]{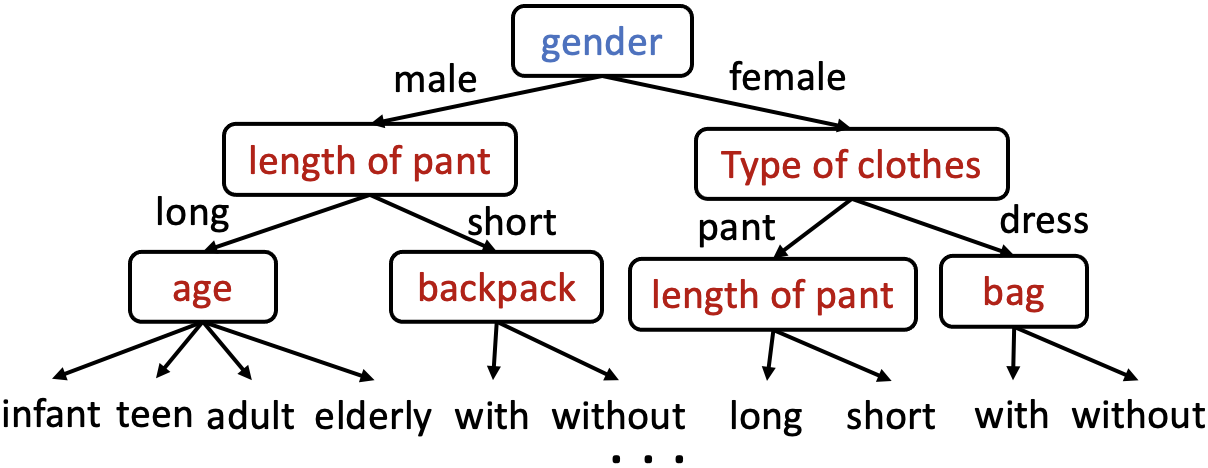}
    \caption{Three levels of our obtained hierarchy to perform person re-identification for the Market-1501 dataset. 
}
    \label{fig:tree}
    
\end{figure}

\vspace{-0.05in}
\subsection{Performing Object Re-Identification}
\label{sec:reID}

The gallery images are first processed by the trained hierarchical DNN. Attributes are assigned to the gallery images by the DNN if the attributes are not available. 
As seen in Fig.~\ref{fig:hier}, a new query image is processed by the root DNN to extract a feature vector and perform an attribute classification to select the next DNN. The feature vector is then processed further by the selected DNN. This process continues until a leaf of the hierarchy is reached. Our experiments use the Euclidean Distance as the distance metric for matching. We only measure the distance between the query image and the gallery images that contain the detected attributes to re-identify the object.
A portion of the obtained hierarchy for the Market-1501 dataset is depicted in Fig.~\ref{fig:tree}. Each image takes a single path from the root to a leaf. Different paths identify different attributes. 

We 
observe that different paths may identify the same attributes (\eg ``length of pant'' in Fig.~\ref{fig:tree}).
Although the corresponding DNNs overlap conceptually, they are specialized to process and identify attributes for different subsets of the gallery, and cannot be used interchangeably. It is important to note that for a query image, only one branch is activated and thus no redundant computation is performed even if the same attributes appear in different branches.



\section{Experiments and Results}
\label{sec:expt}
\vspace{-0.05in}


This section shows that a PyTorch implementation of the proposed method has competitive accuracy, and lower resource requirements than existing techniques. We also show that the method to construct the hierarchy using both visual and semantic similarities is important for the performance gains. The source code is available on GitHub~\cite{source}.

\vspace{-0.05in}
\subsection{Datasets Used}
We use two image datasets: VRAI~\cite{VRAI} for vehicle re-identification, and Market-1501~\cite{Market} for person re-identification. VRAI contains 66,113 images of 6,302 different vehicles.
Market-1501 contains 32,668 images belonging to 1,501 different identities. This dataset also contains 500,000 distractor images for testing. 
Both datasets are divided into training and testing sets and are annotated with attributes~\cite{sem1}. 

\vspace{-0.05in}
\subsection{Experimental Setup}

\textbf{Metrics:} The accuracy is measured using the Rank-1 and Mean Average Precision (mAP) metrics. Rank-1 is the probability that a correct image appears as the top-ranked match. The mAP measures the average retrieval performance when there are multiple matches. 
Accuracy is reported on the testing data. 
The memory requirement and number of operations (FLOPs) for DNNs are found using the \textit{torchsummary} and \textit{thop} PyTorch libraries, respectively. For the proposed method, the worst-case memory requirement and FLOPs are reported: the sum of the model sizes/FLOPs of the DNNs along the longest path from the root to a leaf.
The Yokogawa WT310E Power Meter measures the energy consumption of the techniques on the Raspberry Pi 3 and NVIDIA Jetson Nano.


\textbf{Hierarchical DNN: }
Using the linear evaluation protocol (Section~\ref{subsub:difficulty}) and the correlation matrix (Section~\ref{subsub:corr}), the structure of the hierarchy is defined. The DNNs for each node are then constructed and trained. 
To ensure sufficient training data is available, the depth of a branch is not increased if a child DNN has less than 300 training images.
The deepest branch of the hierarchical DNN is 4 nodes for the VRAI dataset, and 5 nodes for the Market-1501 dataset.
During training, we use the default PyTorch implementation of the triplet loss with batch hard mining. Here, batches are formed by randomly sampling $P$ objects, and then randomly sampling $K$ images of each object, thus resulting in a batch of $P{\cdot}K$ images. We use $P = 8$ and $K = 4$ in our experiments (the largest batch size that fits in available GPU memory). We train with the triplet loss until the loss saturates. The learning rate begins at 0.01 and decays by a factor of 10 every 100 epochs.
When training the DNNs with the cross-entropy loss, we train for 100 epochs with batch size $=$ 32 and learning rate $=$ 0.001. 

\textbf{Comparisons with state-of-the-art:} For the VRAI dataset, RAM-VGG~\cite{RAM}, MultiTask~\cite{VRAI}, and DenseNet201~\cite{Densenet} are used for comparisons. 
For person reID, we compare with pyramidal-reID~\cite{pyramid}, DARE~\cite{wang2018}, Auto-reID~\cite{autoreid}, and PCB~\cite{PCB}. 
These are the state-of-the-art approaches for performing reID.


\textbf{Evaluation of the hierarchy structure:}
We create a \emph{Random Tree} to experimentally show the value of our similarity metric for hierarchy construction. 
In the Random Tree, (1)~the attribute identifications and (2)~their order are selected at random. We expect  the Random Tree to have higher resource consumption and lower accuracy because it does not use the methods described in Section~\ref{sub:order}.
The reported results are based on an average of five runs. For each run, a different random tree is constructed using the method in Section~\ref{sub:cons}. 

\begin{table}[b!]
\centering
\vspace{-0.15in}
\begin{tabular}{|p{0.7cm}|p{2.1cm}|r|r|r|r|}
\hline
 \multicolumn{1}{|c|}{Dataset} & \multicolumn{1}{c|}{Technique} & \multicolumn{1}{c|}{\begin{tabular}[c]{@{}c@{}}Model\\Size\end{tabular}} & \multicolumn{1}{c|}{\begin{tabular}[c]{@{}c@{}}FLOPs\end{tabular}} & \multicolumn{1}{c|}{\begin{tabular}[c]{@{}c@{}}Test\\Rank-1\end{tabular}} & \multicolumn{1}{c|}{\begin{tabular}[c]{@{}c@{}}Test\\ mAP\end{tabular}} \\ \hline
 \multirow{6}{*}{\begin{tabular}[c]{@{}c@{}}VRAI\end{tabular}}
& RAM-VGG~\cite{RAM} & 528 & 15,483 M  &  0.720 &  0.573\\
 & MultiTask~\cite{VRAI} & 103 & 3,882 M &  0.685  &  0.693 \\
 & MultiTask+DP~\cite{VRAI} & 351 & 11,172 M &  \textcolor{blue}{0.803}  & \textcolor{blue}{0.786} \\
 & DenseNet201~\cite{Densenet} & 77 &  4,340 M &   0.671 &  0.700 \\ \cline{2-6}
 &  Random Tree & 25 & 2,026 M &  0.631  & 0.585  \\ \cline{2-6}
 & \textbf{Our Method} &  \textcolor{blue}{14} &  \textcolor{blue}{1,082 M} &  {0.781} & {0.737} \\ \hline \hline 
\multirow{9}{*}{\begin{tabular}[c]{@{}c@{}}Market\\1501\end{tabular}}
& Pyramidal~\cite{pyramid} & 184 & 9,757 M &  0.928  &  0.821 \\
 & DARE~\cite{wang2018} & 89 & 2,891 M &  0.868  &  0.693\\
 & Auto-reID~\cite{autoreid} & 55 & 2,050 M &  \textcolor{blue}{0.938}  & \textcolor{blue}{0.834} \\
 & DG-Net~\cite{DGNet} &  101 &  4,029 M&  0.896 &  0.745\\
 & ResNet50~\cite{resnet} & 103 & 3,882 M &  0.872 & 0.685 \\
 & DenseNet201~\cite{Densenet} & 77 & 4,000 M &   0.860  & 0.699 \\
 & PCB~\cite{PCB} & 107 & 4,206 M &  0.923 &  0.774\\ \cline{2-6}
 &  Random Tree &  27 & 1,736 M &   0.788 & 0.535  \\ \cline{2-6}
& \textbf{Our Method} & \textcolor{blue}{14} &  \textcolor{blue}{808 M} & {0.885} &  {0.699} \\ \hline

\end{tabular}
\caption{Comparison of Model size (in MB), \#operations (FLOPs), rank-1, and mAP. 
Blue font indicates best result.}
\vspace{-0.05in}
\label{Tab:comp}
\end{table}

\vspace{-0.05in}
\subsection{Results}

\begin{figure*}[t!]
	\centering
	\includegraphics[width=0.75\linewidth]{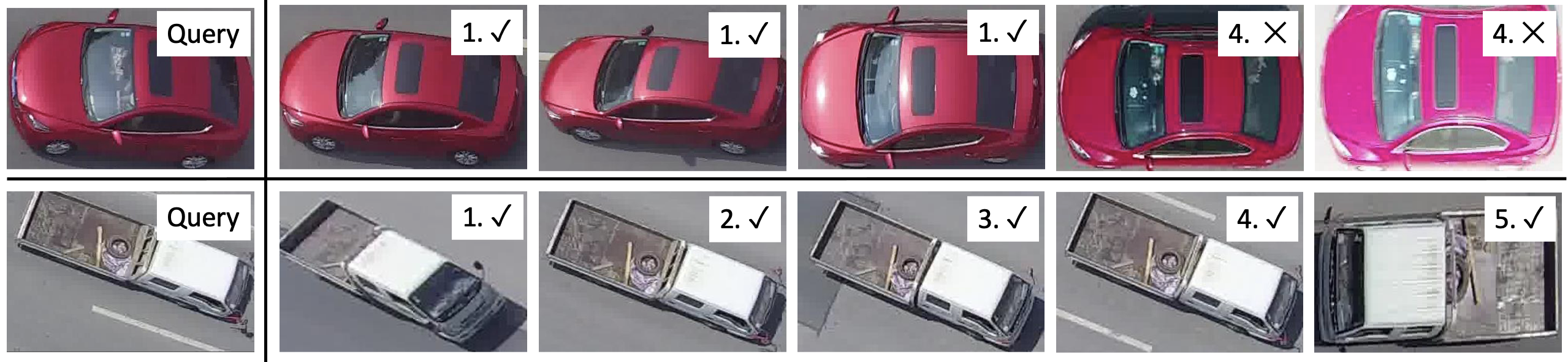}
	\caption{Images returned as matches using the hierarchical DNN for two queries. Correct matches have a check mark.} 
	\label{fig:examples}
	\vspace{-0.15in}
\end{figure*}

TABLE~\ref{Tab:comp} compares the model size, number of operations (FLOPs), rank-1 metric, and mAP of the different techniques. The proposed hierarchical object reID technique requires the least memory. When compared with RAM-VGG, on the VRAI dataset, our method requires 97.3\% ($1 - \frac{14}{528} =$ 0.973) smaller models. 
The FLOPs are also reduced when compared with the state-of-the-art techniques. Our technique requires 80.7\% ($1 - \frac{808}{4206} =$ 0.807)  fewer operations than PCB on the Market-1501 dataset. 
The accuracy of the proposed hierarchical reID system is similar to the state-of-the-art. ResNet50, DenseNet201, and DARE achieve lower accuracy than the proposed technique. The mAP of the proposed technique might improve using random erasing, rank optimization, and image re-ranking~\cite{wang2018}. We do not consider these optimizations.





The comparison with the Random Tree shows that hierarchical DNNs can reduce the resource consumption of object reID. However, random choices are less effective than intelligent hierarchy design.
The Random Tree requires more resources because larger DNNs are used to perform difficult attribute classifications close to the root of the tree. Many redundant attributes are identified and thus deeper trees are required to reduce the gallery size. The average depth of the Random Tree is 6 for VRAI and 7 for Market-1501.
Because the classification error is compounded at every stage of the hierarchy, taller trees achieve lower re-identification accuracy. 


TABLE~\ref{tab:pi} shows the query time and energy consumption of the techniques. The results are reported after averaging over 100 images that use the longest root-leaf path. Our technique requires the least query times and energy on both devices. Techniques that encounter memory errors on the embedded devices are depicted with ``-'' or have been excluded from TABLE~\ref{tab:pi}.
Fig.~\ref{fig:examples} shows reID examples: two query images and the five gallery images that are returned as matches. 
Our method often returns correct matches to the query image.

\begin{table}[]
\centering

\vspace{-0.02in}
\centering
\begin{tabular}{|c|l|r|r|r|r|}
\hline
\multicolumn{1}{|l|}{\multirow{3}{*}{Dataset}} &
  \multicolumn{1}{c|}{\multirow{3}{*}{Technique}} &
  \multicolumn{2}{c|}{\begin{tabular}[c]{@{}c@{}}Raspberry\\ Pi 3\end{tabular}} &
  \multicolumn{2}{c|}{\begin{tabular}[c]{@{}c@{}}NVIDIA Jetson\\ Nano\end{tabular}} \\ \cline{3-6} 
\multicolumn{1}{|l|}{} &
   &
  \multicolumn{1}{c|}{\begin{tabular}[c]{@{}c@{}}Query\\ Time\end{tabular}} &
  \multicolumn{1}{c|}{Energy} &
  \multicolumn{1}{c|}{\begin{tabular}[c]{@{}c@{}}Query\\ Time\end{tabular}} &
  \multicolumn{1}{c|}{Energy} \\ \hline
\multirow{5}{*}{VRAI} & ResNet50~\cite{resnet} & - & - & 3.20 & 22.24 \\ 
                  & DenseNet201~\cite{Densenet} & - & - & 2.75 & 19.26 \\ \cline{2-6}
                  & Random Tree   & 4.85 & 21.99 & 0.78 & 5.68 \\ \cline{2-6}
                &  \textbf{Our Method}  & \textcolor{blue}{2.53} & \textcolor{blue}{12.13} & \textcolor{blue}{0.35} & \textcolor{blue}{2.66} \\ \hline \hline
\multirow{6}{*}{\begin{tabular}[c]{@{}c@{}}Market\\1501\end{tabular}} & ResNet50~\cite{resnet} & - & - & 3.00 & 21.63 \\ 
                  & DenseNet201~\cite{Densenet} & - & - &  2.55 & 18.18  \\ 
                  & DARE~\cite{wang2018} & 11.41 & 55.83 &  1.09 & 7.92   \\ \cline{2-6}
                  & Random Tree  & 4.39 & 19.83 & 0.77 & 5.72 \\ \cline{2-6}
                &  \textbf{Our Method}  &  \textcolor{blue}{2.13} &  \textcolor{blue}{10.33} & \textcolor{blue}{0.35} & \textcolor{blue}{2.70} \\ \hline
\end{tabular}
\caption{Query time (sec/img) and energy  consumption (J/img) comparison on two embedded devices: Raspberry Pi 3 and NVIDIA Jetson Nano. Blue font indicates the best result.}
\label{tab:pi}
\vspace{-0.25in}
\end{table}

\section{Conclusions}
\vspace{-0.05in}

In this paper, we present a novel hierarchical DNN for energy-efficient object re-identification (reID) on embedded devices. 
Our approach employs multiple small DNNs, arranged in the form of a hierarchy, where each DNN processes the input and identifies an attribute. 
Existing hierarchical DNNs use either visual or semantic similarities to construct hierarchies.
Through this work, we demonstrate an approach that takes \emph{both} visual and semantic similarities into account to construct hierarchies for efficient object reID.
This is achieved by quantifying the difficulty of attribute identification and finding the correlation between attributes to determine which attributes are identified, and the order in which they are identified. 
Every time an attribute is identified, we narrow the search space.
We specialize the small DNNs in the hierarchy to process and re-identify only a small subset of the objects to achieve high accuracy with low resource requirements. 
Our experiments confirm that hierarchical DNNs improve the deployability of object reID on two entry-level embedded devices, Raspberry Pi 3 and NVIDIA Jetson Nano. 
We achieve the performance gains due to a systematic approach that constructs an efficient hierarchy, selects DNN sizes, and trains each DNN. In our future work, we will build a framework that allows the exploration of the design space to obtain different trade-offs between accuracy and efficiency.

\section*{Acknowledgement}
\vspace{-0.05in}
This project was supported in part by NSF CNS-1925713. Any opinions, findings, and conclusions or recommendations expressed in this material are those of the authors and do not necessarily reflect the views of the sponsors.

\def\bibfont{\footnotesize}
\printbibliography
\end{document}